\documentclass[twocolumn]{svjour3}          

\usepackage[latin1]{inputenc}
\usepackage{listings}
\usepackage{stmaryrd}
\usepackage{amstext}
\usepackage{amssymb}
\usepackage{dsfont}

\def\TC{Text Classification}
\usepackage{graphicx}
\usepackage{arabtex}
\arabtrue

\begin{document}
\title{Rational Kernels for Arabic Stemming and Text Classification\thanks{This work  is supported by the MESRS - Algeria under Project 8/U03/7015.}}

\author
{Attia Nehar \and Djelloul Ziadi \and Hadda Cherroun
}

\institute{ Attia Nehar \and Hadda Cherroun 
\at Laboratoire d'informatique et Mathématiques\\ 
Université A.T. Laghouat, Algérie\\\email{a.nehar,h.cherroun@mail.lagh-univ.dz}
\and
Djelloul Ziadi 
\at Laboratoire LITIS - EA 4108, \\
Normandie Université, \\
Rouen, France\\\email{djelloul.ziadi@univ-rouen.fr}
}

\authorrunning{Nehar, Ziadi and Cherroun}
\titlerunning{Rational Kernels for Arabic Stemming and \TC}

\maketitle

\begin{abstract}
In this paper, we address the problems of Arabic Text Classification and stemming using Transducers and Rational Kernels. We introduce a new stemming technique based on the use of Arabic patterns (Pattern Based Stemmer). Patterns are modelled using transducers and stemming is done without depending on any dictionary. Using transducers for stemming, documents are transformed into finite state transducers. This document representation allows us to use and explore rational kernels as a framework for Arabic Text Classification. Stemming experiments are conducted on three word collections and classification experiments are done on the Saudi Press Agency dataset. Results show that our approach, when compared with other approaches, is promising specially in terms of Accuracy, Recall and F1.
\end{abstract}

\section{Introduction}\label{sec:intro}
\novocalize
Text Classification (TC) is the task of automatically sorting a set of documents into one or more categories from a predefined set~\cite{Sebas02machinelearn}. Text classification techniques are used in many domains, including mail spam filtering, article indexing, Web searching, automated population of hierarchical catalogues of Web resources, even automated essay grading task. 

Due to the complexity of the Arabic language, Arabic Text Classification (ATC) starts receiving great attention. Many algorithms have been developed to improve performance of ATC systems \cite{alsaleem11,kacst08,duwairi07classi,elkourdi04,gharib2009arabic,hadi08,Kanaan09,khreisatmachine2009,mesleh08,syiam2006}. In general, we can divide an ATC system into three steps:
\begin{enumerate}
	\item \textbf{Preprocessing step}: where punctuation marks, diacritics, stop words and non letters are removed.
	\item \textbf{Features extraction}: a set of features is extracted from the text, which will represent the text in the next step. For instance, Khreisat \cite{khreisatmachine2009} used the N-gram technique to extract features from documents. Another work \cite{syiam2006}, used stemming to extract features.
	\item \textbf{Learning step}: many supervised algorithms were used to learn systems how to classify Arabic text documents: Support Vector Machines \cite{alsaleem11,gharib2009arabic,mesleh08}, K-Nearest Neighbours \cite{hadi08,syiam2006}, Naive Bayes \cite{alsaleem11,elkourdi04} and many others. Most algorithms rely on distance measures over extracted features to decide how much two documents are similar.
\end{enumerate}

In the second step, a feature vector is constructed. Several stemming approaches are developed~\cite{al-nashashibi2010}. Khoja and Garside~(1999)  developed a dictionary based stemmer. It gives good performances, but the dictionary needs to be maintained. The stemmer developed in~\cite{alserhannew2003} finds the three-letter roots for Arabic words without depending on any roots dictionary or pattern files.\\
Many Arabic words have the same stem but not the same meaning. Reducing two semantically different words to the same root can induce classification errors.
To prevent this, light stemming is used in TC algorithms~\cite{Aljlayl02onarabic}. Its main idea is that a lot of words generated from the same root have different meanings. The basis of this light-stemming algorithms consists of several rounds over the text, that attempt to locate and remove the most frequent prefixes and suffixes from the words. This leads to a lot of features due to the light stemming strategy.

In the third step, many distance measures could be used to evaluate distance (or dissimilarity) between documents using these feature vectors. The quality of the classification system is related to the used distance measure.

In this paper, we study the effect of stemming on ATC. Let's illustrate this by an example. Given two simple documents $d_1=$" {\RL{yat`alam wa yataraby al-.tfl fy al-madrasT}" (Child learns and brought up in the school),  and $d_2=$ "{\RL{tqadm al-madArs lA.tfAlnA al-t`alym w al-trbyT}" (Schools provide education for our children). We compute euclidian distance between them using 3-grams, with and without stemming: 

\begin{center}
\begin{tabular}{cc}
\hline\hline
   & \textbf{Distance} \\ 
   &  \textbf{with 3-grams}\\\hline
   	Without stemming		&  0.25 \\ \hline
   	With stemming		  & 0.18  \\ \hline\hline
\end{tabular} 
\end{center}
It is clear that distance between  $d_1$ and $d_2$ is affected by stemming.\\ 

In this work, we enhance the stemming technique, introduced by authors in previous paper \cite{nehar2012a}. Indeed, stemmer introduced in \cite{nehar2012a} gives a set of possible stems. Our new stemmer chooses the best stem based on a statistical study of characters occurences in the Arabic roots corpus. Hence, a comparaison experiment is conducted to assess performances against standard stemmers. This stemming technique transforms documents into finite state transducers. Then, rational kernels~\cite{Cortes04} are used as a framework to do ATC \cite{nehar2013}. This framework enables the use of different distance measures or kernels. 

This paper is organized as follows. Section~\ref{sec:stemming} presents, in more details, the main stemming techniques. In Section~\ref{sec:weighttrans}, we recall some notions on weighted transducers and rational kernels. We present, in Section \ref{sec:framework} our new stemming approach, then we explain how to use rational kernels as a framework for ATC. Experiments and results are reported and interpreted in Section~\ref{sec:experiments}.

\section{Stemming Techniques}\label{sec:stemming}
In the context of ATC, stemming is applied to reduce dimensionality of the feature vectors. Brute stemming (commonly called stemming) transforms each Arabic word in the document, into its root. However, light stemming, reduces word by removing prefixes and suffixes.
\subsection{Brute Stemming} \label{sec:brutestemming}
There are many brute stemming techniques used in the context of ATC. They can be classified into two types: \textit{(i)} \textit{Stemming using a dictionary}, where dictionary of Arabic word stems is needed. \textit{(ii)} \textit{Stemming without dictionary}, where stems are extracted without depending on any root or pattern files.

Khoja and Garside stemmer \cite{khoja99}  removes the longest suffix and the longest prefix. 
It then matches the remaining word with verb and noun patterns, to extract the root by means of a dictionary. The stemmer makes use of many linguistic data files such as a list of all diacritic characters, punctuation characters, definite articles and stop words. This stemmer gives good performance but relies on dictionary which needs to be updated.
The second technique \cite{alserhannew2003}, finds the three-letter roots for Arabic words without depending on any root or pattern files. They extract word roots by assigning weights and ranks to the letters that constitute a word. Consonants were assigned a weight of zero and different weights were assigned to the letters grouped in the word (\RL{s'altmwnyhA}) where all affixes are formed by combinations of these letters. The algorithm selects the letters with the lowest products (weight $ \times$ rank) as root letters. Weights and ranks are assigned to letters using a little bit information on language \cite{alserhannew2003}. This algorithm, like any other brute stemming algorithm, gives the same stem for two semantically different words. 
\subsection{Light Stemming}\label{sec:lightstemming}
In Arabic language, some word variants do not have similar meanings (like the two words:  {\it \RL{mktabT} } which means {\it library} and {\it \RL{kAtb} }which means writer). However, these word variants give the same root if a brute stemming is used. Thus, brute stemming can affect the meaning of words. Light stemming~\cite{Aljlayl02onarabic} aims to enhance the text classification performance while retaining the words meanings. The basis of light-stemming algorithms consists of several rounds over the text, that attempt to locate and remove the most frequent prefixes and suffixes from the word. However, it leads to a lot of features.

\section{Weighted Transducers and Rational Kernels}\label{sec:weighttrans}
Before describing our framework, let's give in what follows, some preliminaries on Weighted Transducers and Rational Kernels.

\textit{Transducers} are finite automata in which each transition is augmented with an output label in addition to the familiar input label. Output labels are concatenated along a path to form an output sequence as with input labels. \textit{Weighted transducers} are finite-state transducers in which each transition carries some weight in addition to the input and output labels. The weight of a pair of input and output strings $(x, y)$ is obtained by summing the weights of the paths labelled with $(x, y)$. The following definition gives a formal definition of weighted transducers~\cite{berstel,Cortes07}. 

\begin{definition}
A weighted finite-state transducer $T$ over the semiring $(\mathds{K}, \oplus ,\otimes , \bar{0}, \bar{1})$ is an 8-tuple:\\ $T=(\Sigma, \Delta, Q, I, F, E, \lambda, \rho)$ where $\Sigma$ is a finite input alphabet, $\Delta$ is a finite output alphabet, $Q$ is a finite set of states, $I \subseteq Q$ the set of initial states, $F \subseteq Q$ the set of final states, $E \subseteq 
Q \times  (\Sigma \cup  \{\epsilon \}) \times (\Delta \cup \{\epsilon \}) \times \mathds{K} \times Q$ a finite set of transitions, $\lambda : I \rightarrow \mathds{K}$ the initial weight function, and $\rho : F \rightarrow \mathds{K}$ the 
final weight function 
\end{definition}

For a path $\pi$ in a transducer, $p[\pi]$ denotes the origin state of that path, $n[\pi]$ its destination state and $w[\pi]$ gives the sum of the weights of its arcs. The set of paths from the initial states $I$ to the final states $F$ labelled with input string $x$ and output string $y$ is denoted by $P(I, x, y, F)$. A transducer $T$ is \textit{regulated} if the output weight associated by $T$ to any pair of input-output strings $(x, y)$ given by:
\begin{eqnarray}
 \llbracket T \rrbracket(x,y)  & = \bigoplus_{\pi \in P(I,x,y,F)} {\lambda(p[\pi]) \otimes w[\pi] \otimes \rho[n[\pi]]}
\end{eqnarray}
is well-defined	in $\mathds{K}$. $\llbracket T \rrbracket(x,y)=\bar{0}$ if $P(I, x, y, F)=~\emptyset$. Figure~\ref{fig:T5} shows an example of a simple transducer, with an input string $x$ : {\it \RL{fA`al} } and an output string $y$ : {\it \RL{f`al} }. The only possible path in this transducer is the singular set : $P(\{0\},x,y,\{4\})$.

Regulated weighted transducers are closed under the following operations called rational operations:
\begin{itemize}
	\item the \textit{sum} (or \textit{union}) of two weighted transducers $T_{1}$ and $T_{2}$ is defined by:
\begin{eqnarray}
\forall(x,y) \in \Sigma^{*} \times \Sigma^{*},\nonumber \\ \llbracket T_{1}\oplus T_{2}\rrbracket (x,y) & = \llbracket T_{1} \rrbracket (x,y) \oplus \llbracket T_{2} \rrbracket (x,y)
\end{eqnarray}
	\item the \textit{product} (or \textit{concatenation}) of two weighted transducers $T_{1}$ and $T_{2}$ is defined by:
\begin{eqnarray}
\forall(x,y) \in \Sigma^{*} \times \Sigma^{*}, \llbracket T_{1} \otimes T_{2} \rrbracket (x,y) =\nonumber \\ \bigoplus_{x=x_{1}x_{2},y=y_{1}y_{2}} \llbracket T_{1} \rrbracket(x_{1},y_{1}) \otimes \llbracket T_{2}\rrbracket(x_{2},y_{2})
\end{eqnarray}

\item The composition of two weighted transducers $T_{1}$ and $T_{2}$ with matching input and output alphabets $\Sigma$, is a weighted transducer denoted by $T_{1} \circ T_{2}$ when the sum:
\begin{eqnarray}\label{eq:compo}
				\llbracket T_{1} \circ T_{2}\rrbracket (x,y) = \bigoplus_{z \ in \Sigma^{*}}{\llbracket T_{1}\rrbracket (x,z)\otimes \llbracket T_{2}\rrbracket (z,y)}
\end{eqnarray}
is well-defined in $\mathds{K}$ for all $x,y \in \Sigma^{*}$
\end{itemize}

Rational Kernels are a general family of kernels, based on weighted transducers, that extend kernel methods to the analysis of variable-length sequences or more generally weighted automata. Let $X$ and $Y$ be non-empty sets. A function $K : X \times Y \rightarrow \mathds{R}$ is said to be a kernel over $X \times Y$. Corinna \textit{et al.}~\cite{Cortes04} give a formal definition for rational kernels:
\begin{definition}
A kernel $K$ over $\Sigma^{*} \times \Delta^{*} $ is said to be rational if there exist a weighted transducer $T=(\Sigma, \Delta, Q,\\ I,F, E, \lambda, \rho) $ over the semiring $\mathds{K}$ and a function $ \varphi : \mathds{K} \rightarrow \mathds{R}$ such that for all $x \in \Sigma^{*}$ and $y \in \Delta^{*}$:
\begin{eqnarray}
	K(x,y) = \varphi(\llbracket T \rrbracket(x,y))
\end{eqnarray}
$K$ is then said to be defined by the pair  $(\varphi,T)$.
\end{definition}

\section{Framework for Arabic Stemming and Text Classification}\label{sec:framework}
In the following we explain how to use transducers to do stemming. First, Arabic patterns, prefixes and suffixes are modelled by simple transducers, then, a stemming transducer is constructed using these simple ones by applying rational operations like concatenation, union and composition. 
Then, we show how to use rational kernels as a framework to do ATC. 
\subsection{Stemming by Transducers}\label{sec:stembytrans}

Arabic language differs from other languages syntactically, morphologically and semantically. One of the main characteristic features is that most words are built up from roots by following certain fixed patterns and adding prefixes and suffixes. For instance, the Arabic word \RL{al-madrasT} (school) is built from the three-letters root or stem \RL{drs} (learn) and using the pattern \RL{mf`al}, then prefix \RL{al-} and suffix \RL{T} (which is used to denote female gender) are added. This results in the measure  \RL{mf`alT} (see Table \ref{root1}). Notice here that the letter \RL{f} denotes the first letter of the three-letters root, \RL{`a} denotes the second letter and \RL{l} denotes the third one.

\noindent We will use measures to construct a transducer which do stemming. Figure~\ref{fig:T5} shows the example of the measure \RL{fA`al}. This transducer ($T_{measure1}$) can be used to extract the three-letters root of any Arabic word matching this measure. This is achieved by composition operation (\ref{eq:compo}).
We consider $T_{word}$, the transducer which maps any string to itself, i.e., the only possible path is the singleton set $P(\{0\},word,word,\{i\})$ (Figure~\ref{fig:almadrasa} shows transducer associated to the Arabic word \RL{mdrsT}).\\
The composition of two transducers is also a transducer.
\begin{eqnarray*}
(T_{word} \circ T_{measure1})(word,y)  &= \\ \sum_{z \in \Sigma^{*}}{T_{word}(word,z) \cdot T_{measure1}(z,y)}
\end{eqnarray*}
Since the only possible string matching $z$ is $z=word$, we conclude that:
\begin{eqnarray*}
(T_{word} \circ T_{measure1})(word,y) & =\\ T_{word}(word,word) \cdot T_{measure1}(word,y)
\end{eqnarray*}
As we have $T_{word}(word,word)= 1$, so: 
\begin{eqnarray*}
(T_{word} \circ T_{measure1})(word,y) & =  T_{measure1}(word,y)
\end{eqnarray*}
If $word$ matches with the measure the output projection will extract the root (or stem) $y$ associated to $word$.

In Arabic language, there are 4 verb prefixes (\RL{n A y t}), 12 noun prefixes (\RL{A, al-, b, t, s, f, lil, l, y, w, n, m})  and more than 20 suffixes (\RL{hlA, tmA, kmA, An, hA, wA, tm, km, tn, kn, nA, tA, mA, wn, yn, hn, hm, th, ty, ny, n, k, h, T, t, A, At, y}}).  \fullvocalize When considering the diacritics, there are more than 3000 patterns
(in our knowledge). Since we don't consider diacritics in our
approach, patterns are much less (less than 200), much of them
are not used in the context of Modern Standard Arabic.
Indeed, the patterns (\RL{fa`ala, fa`ola, fo`aluN, fa`iluN} ) will result in only one pattern \novocalize (\RL{f`al}) after removing diacritics.
For illustration, Tables~\ref{noun:patt},\ref{verb:patt} shows some examples of noun and verb patterns.

We adopt the following process, to construct the stemming transducer, which enable us to include all measures:
\footnotesize
\begin{enumerate}
\item  Building the transducer of all noun prefixes (resp. verb prefixes);
\item  Building the transducer of all noun patterns (resp. verb patterns);
\item  Building the transducer of all noun suffixes (resp. verb suffixes);
\item  Concatenate noun transducers (resp. verb transducers) obtained in 1, 2 and 3.
\item  Sum the two transducers obtained in step 4.
\end{enumerate}
\normalsize
The first and third steps are very simple. We construct a transducer for each prefix (resp. suffix) then we do the union of these transducers. The resulting transducer represents the prefixes (resp. suffixes) transducer (see Figure~\ref{fig:pref} and Figure~\ref{fig:suffixes}). In the second step, we build all possible noun pattern transducers. Then, the sum of these transducers represents the transducer of all noun patterns. We do the same to build the transducer of all verb patterns (Figure \ref{fig:verbs}).  In the forth step, transducers obtained in steps 1, 2 and 3 are concatenated. The final transducer is obtained by the union of transducers built in step 4. 

The resulting transducer $T_{stemmer}$ could not be represented graphically because of large number of states (about 400 states).
This transducer can stem any well-formed Arabic word, i.e, a word which matches with some Arabic measure. In addition, it can give us a semantic information about the stemmed word. This information can be used to improve the quality
of classification system.

Transducers are created and manipulated using the OpenFst library~\cite{openfst}, which is an open source library  for constructing, combining, optimizing, and searching weighted finite-state transducers.

\begin{flushleft}
\textbf{Ponderation of Our Stemmer}
\end{flushleft}
The composition of $T_{stemmer} $ with any given word transducer $T_{word}$ gives a transducer which may include many paths, so many possible roots. Indeed, an Arabic word could match with more than one measure at the same time. Lets take the word \RL{Ant.sr} (win). This Arabic word matches with, at least, two measures: \RL{Anf`al} and \RL{Aft`al} giving the stems \RL{t.sr} and \RL{n.sr} respectively. Thus, the use of $T_{stemmer}$ leads to a set of one or more possible stems. The correct stem belongs to the set of possible stems. To cope with this situation, stemming transducer must be weighted. Many schemes are possible. We use a bigram window probabilities technique to affect a score to a given stem. The technique is based on a statistical study of letter  frequencies in the Arabic roots corpus. This corpus contains more than 10 thousands three letters roots. The score is affected to a given stem by calculating the probability of letter occurrences in different positions. Let $s= c_1c_2c_3$ a three letters stem. $Score(s)$ is calculated by:
$$ Score (s) = P_1(c_1,c_2) \times P_2(c_2,c_3)$$ 
where $P_1(c_1,c_2)$ is the probability to have the letter $c_2$ in the second position preceded by $c_1$, and $P_2(c_2,c_3)$ is the probability to have the letter $c_3$ in the third position preceded by $c_2$. Thus  we consider the correct stem is the one that has the best score $s_{best}$: ${best} = Arg(Max \{Score(s) ~~ s\in \{\text{Possible stems}\}\} )$.

\subsection{Rational Kernels for Arabic \TC}\label{subsec:framework}
Our ATC system is divided into three stages: 
\begin{enumerate}
	\item preprocessing step.
	\item  feature extraction: the previous transducer is applied on each word of the document resulting from step 1. Then, the transducer resulting from the concatenation of these words stems transducers will represent the document in the next step.
	\item learning task: Rational kernels will be used to measure distance between documents~\cite{Cortes04,Cortes07}, and SVM will be used to do classification.
\end{enumerate}

Considering a set of documents, each document consists of a sequence of words: $w_{1}w_{2} \ldots w_{n}$. Applying our stemming transducer on each word of a document and right concatenate results will transform this document into finite state transducer. 
These transducers will be packaged into an archive file (far) to be treated by the learning algorithm (Figure~\ref{fig:trans}). OpenKernel, which is a library for creating, combining and using kernels for machine learning applications, will be used to accelerate experiments.

\section{Experimental Results and Discussion}\label{sec:experiments}

The next batch reports the main commands of OpenFst and OpenKernel libraries used to implement our classification system.

\scriptsize
\begin{lstlisting}
fstcompose word.fst model.fst result.fst
fstconcate doc.fst result.fst doc.fst
farcreate data.list data.far
klngram -order=3 -sigma=29 data.far 3gram.kar
svm-train -k openkernel -K 2gram.kar cul.train cul.train.2gram.mdl
svm-predict cul.test cul.train.2gram.mdl cul.test.2gram.pred
\end{lstlisting}
\normalsize

To stem words in the document, we iterate on these words using the OpenFst command~\cite{openfst} \textit{fstcompose} (line 1), where \textit{word.fst} is a linear finite state transducer with identical input and output labels, which represents a word, and \textit{model.fst} is our ponderated stemming transducer~. The resulting transducer \textit{result.fst} represents the best stem. 
Resulting transducers are right concatenated to a finite state transducer (\textit{doc.fst}), representing the entire document, using the OpenFst command \textit{fstconcate} (line 2).
The set of finite state transducers (FSTs) is then packaged in a FST archive (Far) using the OpenKernel command \textit{farcreate} (line 3), where \textit{data.list} contains the list of all FST documents, one file per line, and \textit{data.far} is the FST archive (Far). 

Various types of kernels could be created using OpenKernel library. 3-gram kernels could be created using the command \textit{klngram} (line 4), where the first argument \textit{--order} specifies the size of the n-grams, and the second argument \textit{--sigma} specifies the size of the alphabet, epsilon not included (Arabic alphabet size is 28). The first parameter is the FST archive (\textit{data.far}) and the second parameter (\textit{3gram.kar}) is the resulting kernel archive.

OpenKernel library includes a plugin for the LibSVM implementation~\cite{libsvm}. This enables us to do training, predicting and scoring on our dataset. Training command creates a model on the training set (line 5), where the first argument -k specifies the kernel format, the second one (-K) specifies the n-gram kernel archive. The first parameter specifies a correctly classified subset of the training set, the second parameter is the resulting model. In this command, \textit{cul.train} contains a labelled sub set of training documents belonging to Cultural class. Having a model, we can use it to classify documents of the testing dataset with the command \textit{svm-predict} (line 6), where the first parameter specifies a correctly classified subset of the testing set, the second parameter is the resulting model from the previous command. The last parameter contains the result of prediction using the model.

\subsection{Stemming Results}
To check the performances of our stemmer, experiments were performed on three word collections. The first one (Gold1) is a sample taken from the Corpus of Contemporary Arabic \cite{majdiopen-source2011}. The two others (Gold2 and Gold3) are house built sets. All words of these sets were annotated by hand with the correct root. Roots have been checked by Arabic Language scholars who are experts in the Arabic Language. The three sets are picked randomly from different topics, including politics, culture, sport and news. Table~\ref{tab:words} gives an overview of these three collections. We give for each gold, the number of words (\# words).
Table~\ref{tab:stem} reports the accuracy of our stemmer on the three sets of words.

Experiment results show the effectiveness of our approach of stemming. Results on different corpora are stable and the best score is achieved with the greatest corpus (Gold3). 
Our stemmer results are sandwiched between Khoja and Al-Serhan stemmer results. This can be explained by the fact that Khoja's stemmer is a dictionary based tool, which makes it language dependent. Al-Serhan stemmer is an unsupervised one. It uses a little bit information about the language. Our stemmer is a semi-supervised tool. It uses a language knowledge -patterns- but only in the construction stage. Patterns are fixed and do not change.

\subsection{ATC Results}

We perform experiments on the Saudi Press Agency (SPA) dataset~\cite{kacst08} for training and testing the  ATC system. As detailed on Table~\ref{tab:spa}, this dataset contains 1,526 text documents belonging to one of the six categories: culture, economic, social, general, politics and sport. As mentioned before, stop words, non Arabic letters, symbols and digits were removed. We have used $80\%$ of documents for training the classifier and $20\%$ for testing. Learning is done using LibSVM implementation~\cite{libsvm}, included in Openkernel, with three different n-gram kernels $(n=2,3,4)$. Since we want to show the effect of stemming, we report results of the three classifier versions; without stemming (Classifier 1), with Al-Serhan  stemmer (Classifier 2) and with our stemmer (Classifier 3), in terms of accuracy, precision, recall and F1.
In Figures~\ref{fig:2gramA}, \ref{fig:3gramA} and \ref{fig:4gramA}, we report results in terms of accuracy and precision for the three classifiers with the three kernels (bigrams, 3-grams and 4-grams). Figures~\ref{fig:2gramB}, \ref{fig:3gramB} and \ref{fig:4gramB} give results in terms of recall and F1 for the same classifiers.


Concerning the quality of classification, Figure~\ref{fig:accprec} shows that
best results were reached with 3-grams kernel for accuracy, recall and
F1 measures. This can be explained by the fact that over than 80\% of
Arabic words are built from 3-letter roots.
   
For the 3-gram kernel, let us measure the effect of stemming on classification. For most classes, stemming enhance results in terms of accuracy, Recall and F1 (see Figures \ref{fig:3gramA} and \ref{fig:3gramB}). However, for precision, stemming affects negatively performances (see Figure \ref{fig:3gramA}). 

One can argue the best scores observed by sport class by the fact that it uses a specific vocabulary. Poor results are reported for the General class. This is expected given the used words in this kind of documents which are generic. 
At last, our classifier surpasses other classifiers in most cases. 


\section{Conclusion} \label{sec:conc}
In this paper we introduced a new framework for Arabic word stemming and Text classification. It is based on the use of transducers for stemming, and rational kernels for measuring distance between documents. First, our stemmer uses transducers for modelling Arabic patterns. Second, rational kernels are used to measure distances between documents.
Experiments and analysis of this framework in the context of Arabic Text Classification show that stemming improves the quality of classifiers in terms of accuracy, recall and F1. But it lightly decreases the precision. 3-grams based classifiers reached the best results. Like that of Al-Serhan, our approach of stemming do not rely on dictionary, and it gives better results.  

In future work, other kernels, like word-grams and gappy grams, will be investigated. 

\newpage

\begin{table*}[!htb]
\caption{Measures for the 3-letters root \RL{d r s} and built words.}\label{root1}
\scriptsize
\begin{tabular}{lccccc}
\hline\hline
\textbf{Measures} &  { \RL{mf`alaT} }& { \RL{fA`al}  } & {\RL{al-f`aAlT} }& { \RL{yf`al} }& { \RL{ytfA`al} } \\ 
\hline
\textbf{Words} & {\it \RL{mdrsaT} }		     &    {\it \RL{dArs} }   &    {\it \RL{al-drAsT} }		& {\it \RL{ydrs} }		  &  {\it \RL{ytdArs} }	 \\
\hline\hline
\end{tabular} 
\normalsize
\end{table*}

\begin{table*}[!htb]
\caption{Examples of noun patterns.}\label{noun:patt}
\scriptsize
\begin{tabular}{ccccc}
\hline\hline
 \multicolumn{5}{c}{\it Noun Patterns }\\ \cline{1-5} 
\textbf{3-letters} &\textbf{4-letters} & \textbf{5-letters} &\textbf{6-letters} &\textbf{7-letters} \\
\hline
\RL{fa`ala}   & \RL{fA`ala} & \RL{mfA`l} & \RL{mtfA`al} & \RL{Astf`Al} \\ 
   				 & \RL{fa`wl} & \RL{mft`al} & \RL{mf`aw`l} & \RL{Af`ylAl} \\ 
     			 &	\RL{mf`l}& \RL{mft`l} & \RL{mstf`l} & \RL{Aft`AlT} \\ 
\hline\hline
\end{tabular} 
\normalsize
\end{table*}

\begin{table*}[!htb]
\caption{Examples of verb patterns.}\label{verb:patt}
\scriptsize
\begin{tabular}{ccc}
\hline\hline
 \multicolumn{3}{c}{\it Verb Patterns} \\ \cline{1-3} 
 \textbf{3-letters} &\textbf{4-letters} & \textbf{3-letters +1} \\
\hline
\RL{f`al}& \RL{fa`lala} &\RL{fA`ala} \\ 
\cline{1-3} 
\textbf{3-letters +2} &\textbf{3-letters +3} &\textbf{4-letters +1} \\
\hline
\RL{Afta`ala}   & \RL{Astaf`ala} & \RL{tafa`lala}\\
\RL{Anfa`ala}   & \RL{Af`awlala} & \RL{Af`anlala} \\ 
\RL{tafA`la}   & 						&  \\ 
\hline\hline
\end{tabular} 
\normalsize
\end{table*}

\begin{table*}[!htb]
\caption{Gold Standards details.}\label{tab:words}
\scriptsize
\begin{tabular}{lc}
\hline
\textbf{Corpus}  & \# \textbf{words} \\\hline
   	Gold1		& 679 \\ \hline
   	Gold2		& 844  \\ \hline
   	Gold3		& 1,000\\ \hline
\end{tabular} 
\normalsize
\end{table*}

\begin{table*}[!htb]
\caption{Accuracy of Stemmers.}\label{tab:stem}
\scriptsize
\begin{tabular}{lcccc}
\hline
\textbf{Corpus} & \textbf{Khoja}  & \textbf{Our} & \textbf{Al-Serhan}\\ 
				& \textbf{Stemmer}  & \textbf{Stemmer} & \textbf{Stemmer}\\ 
				& \textbf{\% }  & \textbf{\% } & \textbf{\%}\\ \hline
   	\textbf{Gold1}	& 82,77	&  71.68 & 51,40 \\ \hline
   	\textbf{Gold2}	& 85,55	&  74,82 & 49,64 \\ \hline
   	\textbf{Gold3}	& 87,60	&  80.30 & 56,40 \\ \hline
   	\textbf{Average} & 85,30  & 75.60 & 52,48 \\ \hline
\end{tabular} 
\normalsize
\end{table*}

\begin{table*}[!htb]
\caption{SPA corpus details.}\label{tab:spa}
\scriptsize
\begin{tabular}{cccc}
\hline
\textbf{Categories}   & \textbf{Training texts} & \textbf{Testing texts} & \textbf{Total} \\ \hline
   	Culture			 & 201 & 57 & 258\\ 
   	Economics		 & 200 & 50 & 250\\ 
   	Social			 & 203 & 55 & 258\\ 
   	Politics		 & 200 & 50 & 250\\ 
   	General			 & 205 & 50 & 255\\ 
   	Sports			 & 205 & 50 & 255\\ \hline
   							 & 1,214& 312& 1,526\\ \hline
\end{tabular} 
\normalsize
\end{table*}


\begin{figure}[!htb]
	\centering
		\includegraphics[width=0.50\textwidth]{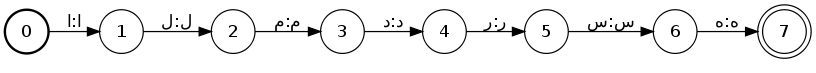}
	\caption{Transducer corresponding to the word \RL{al-mdrsT} (school).}
	\label{fig:almadrasa}
\end{figure}

\begin{figure}[!htb]
	\centering
\vspace{3cm}
		\includegraphics[width=0.5\textwidth]{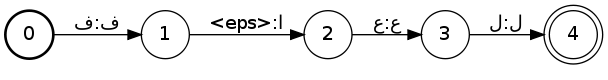}
	\caption{Example of a transducer.}
	\label{fig:T5}
\end{figure}

\begin{figure}[!htb]
	\centering
		\includegraphics[width=0.35\textwidth]{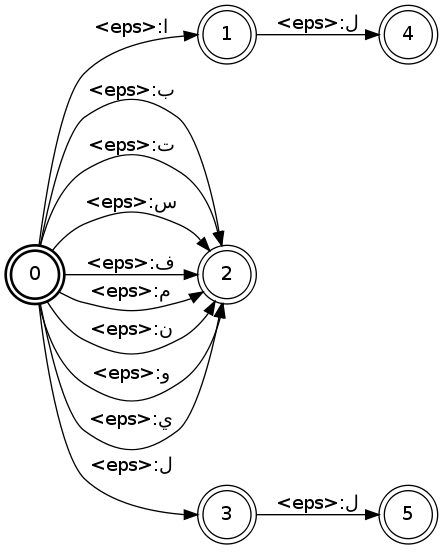}
		\includegraphics[width=0.2\textwidth]{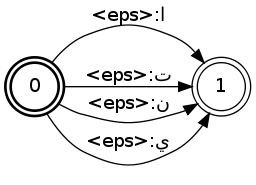}
	\caption{Transducer of noun prefixes (top) and verb prefixes (bottom).}
	\label{fig:pref}
\end{figure}

\begin{figure}[!htb]
	\centering
		\includegraphics[width=0.45\textwidth]{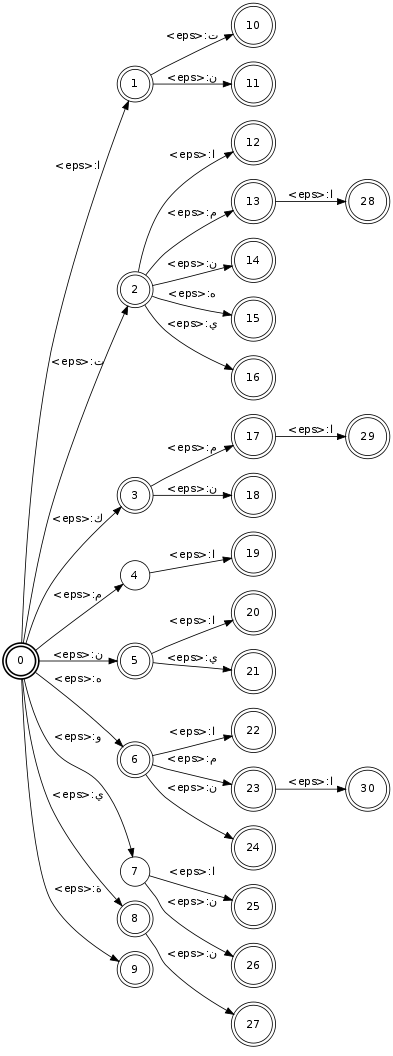}
	\caption{Transducer of noun and verb suffixes.}
	\label{fig:suffixes}
\end{figure}

\begin{figure}[!htb]
	\centering
		\includegraphics[width=0.45\textwidth]{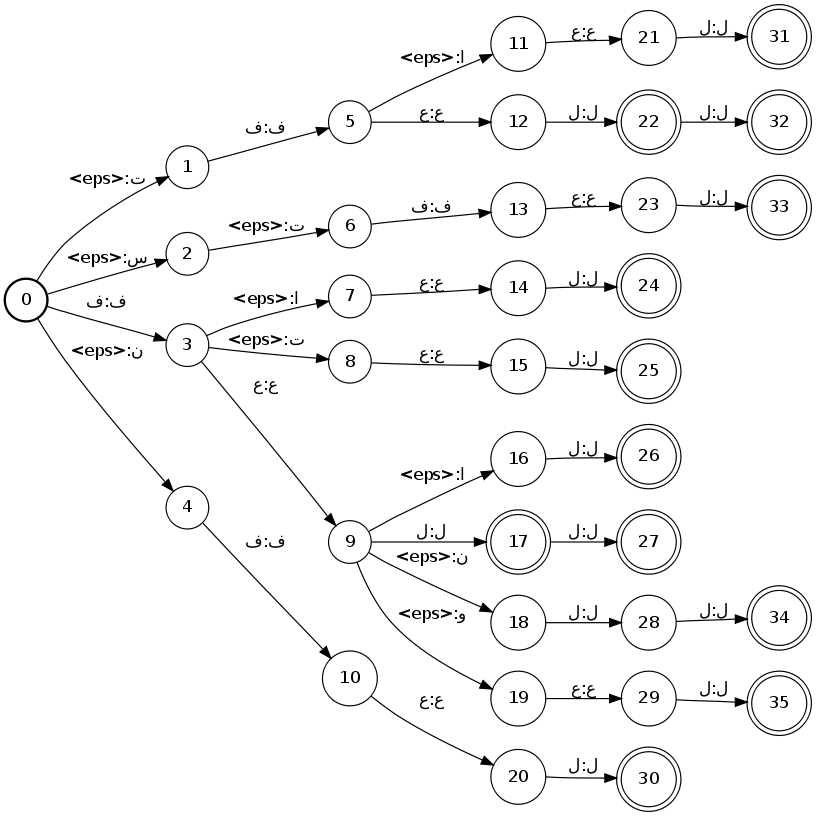}
	\caption{Transducer of verb patterns.}
	\label{fig:verbs}
\end{figure}

\begin{figure}[!htb]
	\centering
		\includegraphics[width=0.4\textwidth]{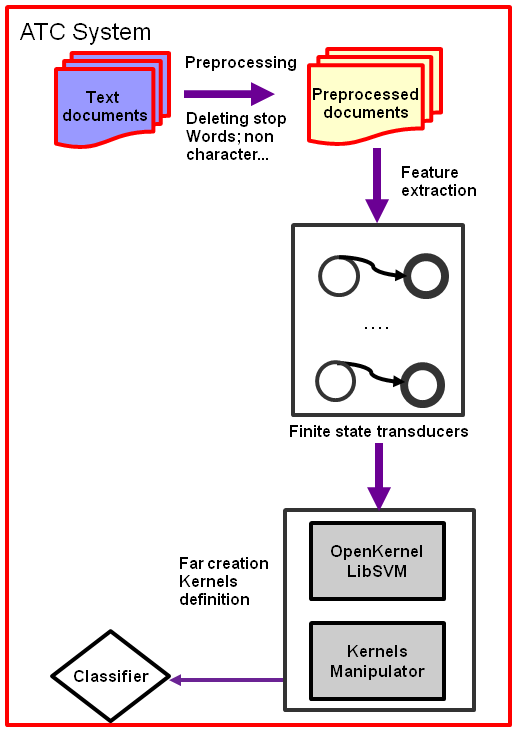}
	\caption{Transformation of a text document into a finite state transducer.}
	\label{fig:trans}
\end{figure}

\begin{figure}[!htb]
\includegraphics[width=0.5\textwidth]{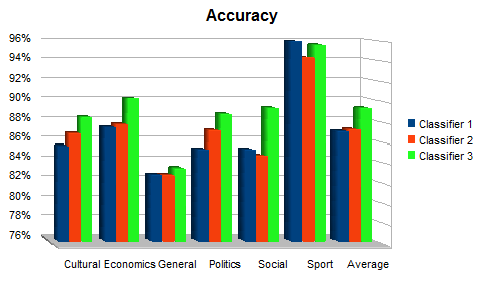}
\includegraphics[width=0.50\textwidth]{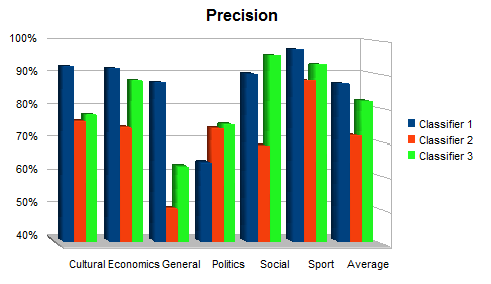}
\caption{Accuracy and Precision of SVM Classification using 2-gram Kernel.}
	\label{fig:2gramA}
\end{figure}

\begin{figure}[!htb]
\includegraphics[width=0.5\textwidth]{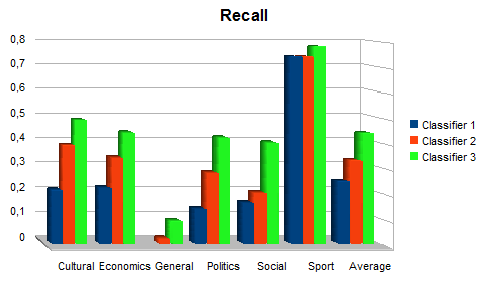}
\includegraphics[width=0.5\textwidth]{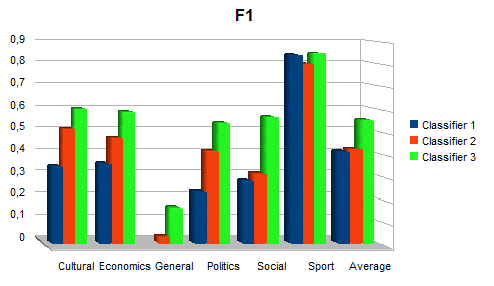}
\caption{Recall and F1 of SVM Classification using 2-gram Kernel.}
	\label{fig:2gramB}
\end{figure}

\begin{figure}[!htb]
\includegraphics[width=0.5\textwidth]{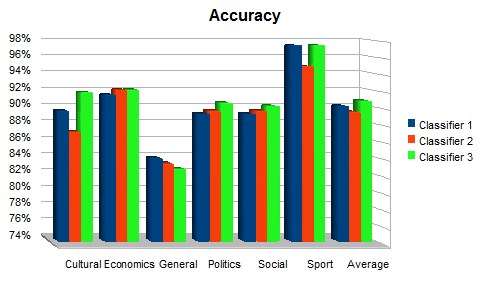}
\includegraphics[width=0.50\textwidth]{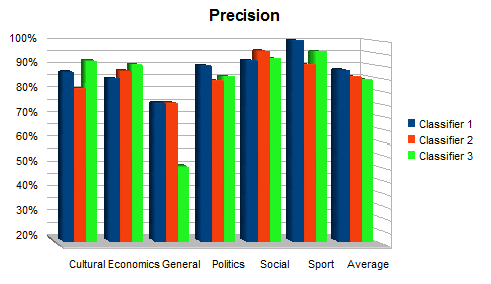}
\caption{Accuracy and Precision of SVM Classification using 3-gram Kernel.}
	\label{fig:3gramA}
\end{figure}

\begin{figure}[!htb]
\includegraphics[width=0.5\textwidth]{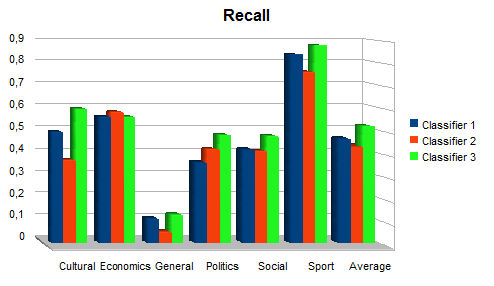}
\includegraphics[width=0.5\textwidth]{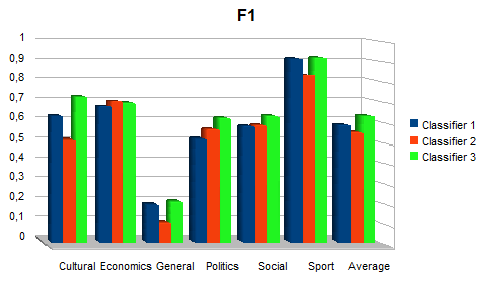}
\caption{Recall and F1 of SVM Classification using 3-gram Kernel.}
	\label{fig:3gramB}
\end{figure}

\begin{figure}[!htb]
\includegraphics[width=0.5\textwidth]{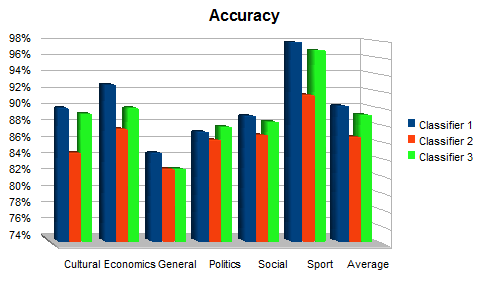}
\includegraphics[width=0.50\textwidth]{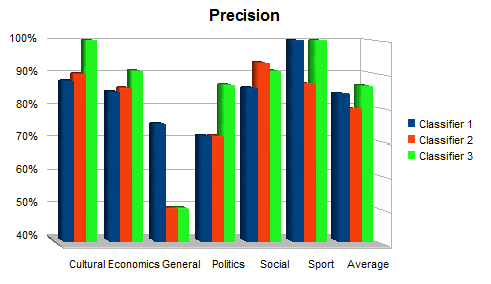}
\caption{Accuracy and Precision of SVM Classification using 4-gram Kernel.}
	\label{fig:4gramA}
\end{figure}

\begin{figure}[!htb]
\includegraphics[width=0.5\textwidth]{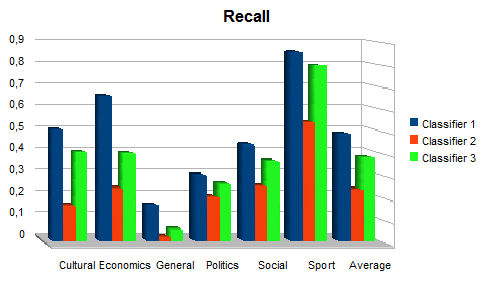}
\includegraphics[width=0.5\textwidth]{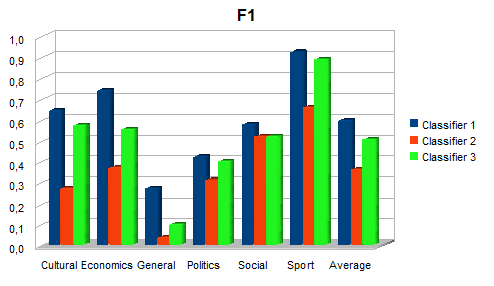}
\caption{Recall and F1 of SVM Classification using 4-gram Kernel.}
	\label{fig:4gramB}
\end{figure}

\clearpage
\begin{figure}[!htb]
\includegraphics[width=0.5\textwidth]{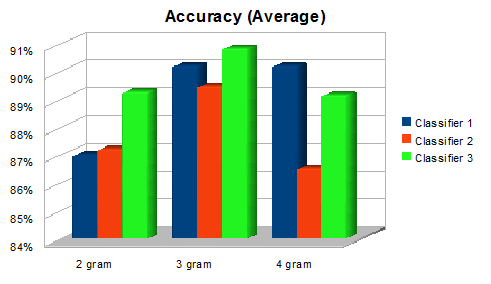}
\includegraphics[width=0.5\textwidth]{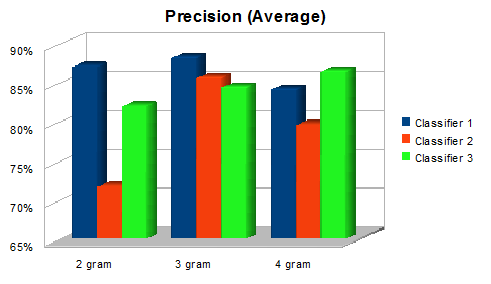}
\caption{Accuracy and Precision averages using Bigram, 3-gram and 4-gram Kernels.}
	\label{fig:accprec}
\end{figure}

\bibliographystyle{spmpsci}
\clearpage

\bibliography{Mesreference}

\end{document}